\DocumentMetadata{}
\PassOptionsToPackage{prologue,dvipsnames,svgnames}{xcolor}
\documentclass[sigconf]{acmart}
\usepackage{graphicx}

\usepackage{subcaption}
\usepackage{multirow}
\usepackage{array}
\usepackage{bm}
\usepackage{enumitem}

\DeclareMathOperator*{\argmin}{argmin} 
\DeclareMathOperator*{\argmax}{argmax} 
\usepackage{tabularx, booktabs}
\usepackage{pgfplots}
\usepackage{tikz}
\usepackage{xcolor}
\usepackage{tablefootnote}
\usepackage{threeparttable}
\usepackage{amsmath,amsthm,mathtools}
\usepackage{dsfont}
\newcommand{\tabincell}[2]{\begin{tabular}{@{}#1@{}}#2\end{tabular}}
\pgfplotsset{compat=1.11,
    /pgfplots/ybar legend/.style={
    /pgfplots/legend image code/.code={%
       \draw[##1,/tikz/.cd,yshift=-0.25em]
        (0cm,0cm) rectangle (3pt,0.8em);},
   },
}
\definecolor{applegreen}{rgb}{0.55,0.71,0.0}

\AtBeginDocument{%
  \providecommand\BibTeX{{%
    \normalfont B\kern-0.5em{\scshape i\kern-0.25em b}\kern-0.8em\TeX}}}

\copyrightyear{2024}
\acmYear{2024}
\setcopyright{rightsretained}
\acmConference[CIKM '24] {Proceedings of the 33rd ACM International Conference on Information and Knowledge Management}{October 21--25, 2024}{Boise, ID, USA.}
\acmBooktitle{Proceedings of the 33rd ACM International Conference on Information and Knowledge Management (CIKM '24), October 21--25, 2024, Boise, ID, USA}
\acmISBN{979-8-4007-0436-9/24/10}
\acmDOI{10.1145/3627673.3679722}

\newtheorem{theorem}{Theorem}

\begin{document}

\title{Retrieval-enhanced Knowledge Editing in Language Models for Multi-Hop Question Answering}

\author{Yucheng Shi}
\orcid{0009-0007-4192-1315}
\affiliation{%
  \institution{University of Georgia}
  \city{Athens}
  \state{Georgia}
  \country{USA}
}
\email{yucheng.shi@uga.edu}

\author{Qiaoyu Tan}
\orcid{0000-0001-8999-968X}
\affiliation{%
  \institution{New York University}
  \city{New York}
  \country{USA}
}
\email{qiaoyu.tan@nyu.edu}

\author{Xuansheng Wu}
\orcid{0000-0002-7816-7658}
\affiliation{%
  \institution{University of Georgia}
  \city{Athens}
  \state{Georgia}
  \country{USA}
}
\email{xuansheng.wu@uga.edu}

\author{Shaochen Zhong}
\orcid{0009-0001-7289-3667}
\affiliation{%
  \institution{Rice University}
  \city{Houston}
  \state{Texas}
  \country{USA}
}
\email{shaochen.zhong@rice.edu}

\author{Kaixiong Zhou}
\orcid{0000-0001-5226-8736}
\affiliation{%
  \institution{North Carolina State University}
  \city{Raleigh}
  \state{North Carolina}
  \country{USA}
}
\email{kzhou22@ncsu.edu}

\author{Ninghao Liu}
\orcid{0000-0002-9170-2424}
\affiliation{%
  \institution{University of Georgia}
  \city{Athens}
  \state{Georgia}
  \country{USA}
  }
\email{ninghao.liu@uga.edu}

\renewcommand{\shortauthors}{Yucheng Shi, et al.}
\newtheorem{problem}{Problem}

\begin{abstract}
Large Language Models (LLMs) have shown proficiency in question-answering tasks but often struggle to integrate real-time knowledge, leading to potentially outdated or inaccurate responses. This problem becomes even more challenging when dealing with multi-hop questions, since they require LLMs to update and integrate multiple knowledge pieces relevant to the questions. To tackle the problem, we propose the Retrieval-Augmented model Editing (RAE) framework for multi-hop question answering. RAE first retrieves edited facts and then refines the language model through in-context learning. Specifically, our retrieval approach, based on mutual information maximization, leverages the reasoning abilities of LLMs to identify chain facts that traditional similarity-based searches might miss. In addition, our framework includes a pruning strategy to eliminate redundant information from the retrieved facts, which enhances the editing accuracy and mitigates the hallucination problem. Our framework is supported by theoretical justification for its fact retrieval efficacy. Finally, comprehensive evaluation across various LLMs validates RAE's ability in providing accurate answers with updated knowledge. Our code is available at: \url{https://github.com/sycny/RAE}.
\end{abstract}




\maketitle

\section{Introduction}

\begin{figure}
    \centering
    \includegraphics[width=0.975\linewidth]{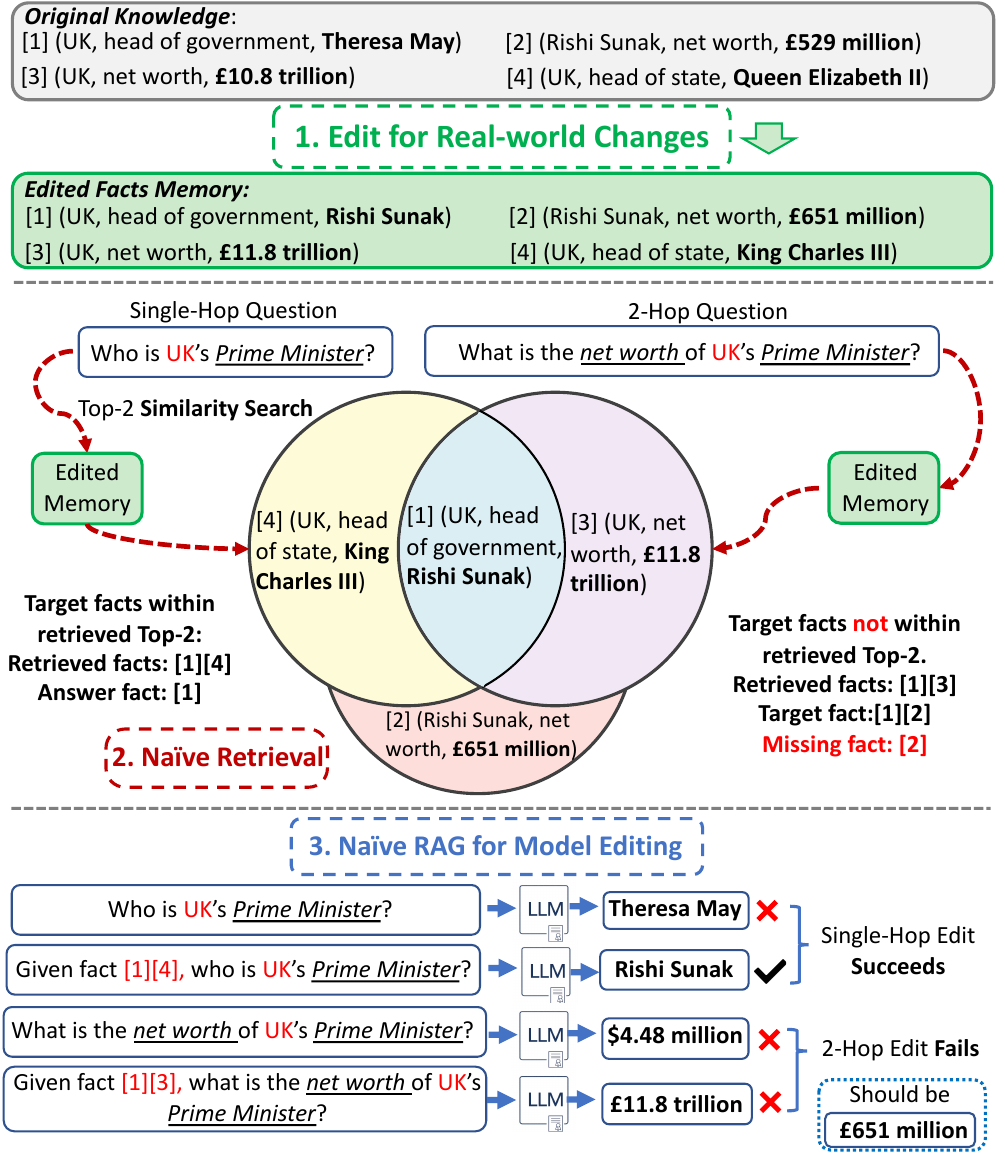}
    \caption{An example of traditional similarity-based search that fails to retrieve the correct facts for LLM editing.}
    \label{intro}
    \vspace{-15pt}
\end{figure}
Large language models (LLMs) excel in answering factual questions due to their pre-training on extensive corpora~\cite{radford2019language, gpt-j, touvron2023llama}. However, their dependence on pre-trained knowledge can lead to outdated answers, which requires updates through model editing techniques~\cite{meng2022locating, yao2023editing, zhang2024comprehensive, zhong2024gnns}. Among them, editing for \textbf{multi-hop questions} is particularly important, as real-world questions often require combining multiple knowledge pieces. For example, to answer "\textit{Who is married to the British Prime Minister?}", one must connect several related facts, creating a \textbf{fact chain}, like "\textit{(United Kingdom, head of government, \colorbox{LightPink}{Theresa May})}, "\textit{(\colorbox{LightPink}{Theresa May}, spouse, Philip May)}". If we update "\textit{\colorbox{LightPink}{Theresa May}}" to "\textit{\colorbox{LightSkyBlue}{Rishi Sunak}}" to reflect real-world change~\cite{wiki:rishisunak}, it requires adjusting the related facts accordingly, resulting in a new fact chain: "\textit{(United Kingdom, head of government, \colorbox{LightSkyBlue}{Rishi Sunak})}, \textit{(\colorbox{LightSkyBlue}{Rishi Sunak}, spouse, Akshata Murty)}". 
Recent research shows that retrieval-augmented generation (RAG) is effective in LLM editing for multi-hop question answering~\cite{zhong2023mquake, wang2024deepedit}, surpassing other methods such as fine-tuning~\cite{zhu2020modifying} and locate-and-edit~\cite{meng2022locating, meng2022mass} approaches. The RAG-based methods first retrieve relevant facts related to the question and then feed these facts into LLMs through in-context learning. Studies have shown that LLMs are highly receptive to external knowledge, even when it contradicts their internal pre-trained knowledge~\cite{xie2023adaptive, kortukov2024studying}. Therefore, RAG can effectively update knowledge within LLMs but also avoid problems such as catastrophic forgetting~\cite{gu2024model, gupta2024model} and hallucinations~\cite{gekhman2024does}. However, simply using multi-hop questions as queries in RAG often fails to retrieve pertinent facts due to the complexity of the questions involved, as illustrated in Figure~\ref{intro}.

To address this, some methods~\cite{wang2024deepedit, zhong2023mquake} propose breaking down complex multi-hop questions into simpler, single-hop queries to improve the effectiveness of the retrieval process. Despite enhancements, critical issues persist: 1) Commonly used models with fewer parameters (e.g., Llama2-7B~\cite{touvron2023llama2}) exhibit significantly worse editing performance (more than 30\% gap) compared to more powerful models like GPT-3.5 or GPT-4~\cite{zhong2023mquake}; 
2) The performance of existing editing methods degrades as the number of editing cases increases, making them impractical for large-scale editing tasks~\cite{zhong2023mquake, wang2024deepedit}. The possible reason behind their performance degradation is that their editing effectiveness heavily relies on the quality of sub-questions generated for querying. Generating accurate sub-questions is challenging for models with less strong reasoning and planning capabilities. Inaccurate sub-questions will result in irrelevant fact retrieval, which may mislead the LLMs and reduce editing effectiveness~\cite{yoran2023making}. 

Since successful multi-hop editing depends on \textbf{accurate retrieval of the question-specific facts}, instead of applying the existing "generate then retrieve" approach, 
we propose to directly fetch the required facts from the database by leveraging the next-token prediction capabilities of LLMs. 
Below are the details of our editing method.

\textbf{1. The connection between edited facts helps retrieval}: Each edited fact can be represented as a triplet of \textit{"(head entity, relation, tail entity)"}. The nearby facts in a fact chain are connected through entities: the tail entity of one fact becomes the head entity of the next. This observation inspires us to adopt a knowledge graph (KG) for storing these triplets. In a KG, each entity has only a limited number of neighboring entities, which narrows down the retrieval choices to a feasible number, rather than having to search through a vast database. Our method differs from existing approaches that store edited facts as embeddings in a vector database, whose editing performance decreases as the number of edits grows~\cite{zhong2023mquake, wang2024deepedit}.

\textbf{2. Next-token prediction helps next-fact retrieval}: LLMs are inherently skilled at predicting the next token in a sequence. In our study, we extend this capability to predict the next fact in a fact chain. Our approach first feeds LLMs a sequence that includes the question, any preceding fact, and the relevant entity. Then, the model predicts this entity's next possible logical relation within the context of its input question. Finally, we retrieve the tail entity from the KG, based on the head entity and relation, to complete the fact chain. However, directly predicting the most probable next fact can lead to biases toward frequently occurring words~\cite{welbl2021challenges, sun2019mitigating}. Therefore, we predict facts that share the most relevant information with the question and preceding facts, using mutual information (MI) as our retrieval metric. We develop a technique to break down the MI into conditional probabilities that LLMs can effectively approximate~\cite{radford2019language}, thus improving the accuracy of our predictions.

\textbf{3. LLM internal state helps reduce redundancy}: Irrelevant facts to the editing will mislead LLMs. Thus, we propose to prune redundant facts extracted by the retrieval step using the LLM's prediction uncertainty. The uncertainty is minimized when the LLM is prompted with a correct fact chain, and increases when provided with incomplete or excessive facts. We quantify this uncertainty using the LLM's output entropy. Unlike traditional methods that either limit the number of retrieval attempts or simply ask the model to identify redundant facts~\cite{zhong2023mquake, wang2024deepedit}, our approach provides a more effective way to ensure accuracy in the editing process.

Overall, we name our approach as \underline{R}etrieval-\underline{A}ugmented model \underline{E}diting (RAE), where we introduce a novel fact retrieval method for multi-hop questions in model editing. We also propose a knowledge-pruning strategy to reduce noise after the initial retrieval, mitigating the hallucination problem. Additionally, we provide theoretical analysis to justify our design for the retrieval objective.

\section{Preliminary: Model Editing}
\subsection{Model Editing for Single-hop Questions}
In LLMs, \textit{a single model edit} refers to updating a specific piece of factual knowledge~\cite{mitchell2021fast, meng2022mass, zheng2023can}. Each \textit{knowledge} is defined as a triplet $\delta := (h, r, t)$, where $h$, $r$, and $t$ denote the head entity, the relation, and the tail entity, respectively, such as (Misery, author, Stephen King ).
An edit is defined as changing the tail entity $t$ to a new entity $t'$, i.e., $\delta \rightarrow \delta' := (h, r, t) \rightarrow (h, r, t')$, where $\delta'$ is the edited knowledge.   
Let $q$ denote the language model's input. 
The goal of model editing is to modify a target model $f_{\theta}$, so that the new model $f_{\theta}'$ produces an output $f_{\theta}'(q)$ that follows the new fact $\delta'$, where $f_{\theta}'(q)\neq f_{\theta}(q)$. 
Specifically, given $q=[h;r]$, $h,r\in \delta'$, the model is expected to output $t'=f_\theta'([h;r])$, where $[;]$ is the concatenation operator. 
However, if the input question is not relevant to the edit, i.e., $h \notin \delta'$ or $r \notin \delta'$, the model should output $t = f_{\theta}'([h;r])$ that reflects the original knowledge of LLMs.

\subsection{Model Editing for Multi-hop Questions}
Answering multi-hop questions presents a greater challenge. A multi-hop question seeks to identify a specific tail entity $t_k$ based on a sequence of linked facts: $\{(h_1,r_1,t_1), (h_2,r_2,t_2), ..., (h_k,r_k,t_k)\}$, where each tail entity is the head entity of the next fact: $t_i = h_{i+1}$. 
Answering each input question $q$ requires a \textbf{fact chain} $G_q$.
A $k$-hop question can be formulated using only the initial head entity $h_1$, and a series of relationships $\{r_1, r_2, ..., r_k\}$. 
An example of model editing for a 3-hop question is shown in Table~\ref{table1}. Here, we use counterfactual edits to simulate real-world updates. Different fact chains are used to answer the question before and after editing.
One key observation is that fact chains represent connected knowledge graphs, where a single entity is involved in two consecutive facts. Additionally, we notice a "\textbf{ripple effect}" in these chains: An edit in the first fact  $\delta_1$ will lead to changes in the subsequent facts, forming a new chain $G^*_q$.
In practice, knowledge editing is usually conducted in batches, involving multiple fact changes simultaneously, resulting in an edited fact bank $\Delta = \{\delta_1', \delta_2',...,\delta_N'\}$, where $N$ is a large number~\cite{meng2022mass}. 
Locating the relevant edited facts for a question is non-trivial due to the "ripple effect". To correctly answer multi-hop questions after model editing, it is crucial to address the retrieval problem formally defined as follows:
\begin{table}[t]
\centering
\small
\caption{Answering a 3-hop question $q$ with counterfactual model editing. The pre-edited and edited answer are $t_5$ and $ t_3^*$, respectively. $t_5$ and $ t_3^*$ are the tail entity of $\delta_5$ and $\delta'_3$. $G_q$ and $ G^*_q$ denote the pre-edited and edited fact chain.}
\label{table1}
\begin{tabularx}{\columnwidth}{X}
\toprule
\textbf{Edited Fact Bank $\Delta$ and Unedited Facts}   \\
$(\delta_1\rightarrow\delta_1')$ \colorbox{LightGrey}{Misery, author, Stephen King $\rightarrow$ Richard Dawkins} \, \color{gray}{\#edit}  \\
$(\delta_2)$ Richard Dawkins, citizen of, United Kingdom \\
$(\delta_3 \rightarrow \delta_3')$ \colorbox{LightGrey}{United Kingdom, capital, London $\rightarrow$  Birmingham} \,\, \color{gray}{\#edit} \\
$(\delta_4)$ Stephen King, citizen of, United States\\
$(\delta_5)$ United States, capital, Washington, D.C. \\
\midrule
\textbf{A 3-hop Question} $q$ \\
$(q)$ Which city is the capital of the country where the author of \colorbox{LightSkyBlue}{Misery} held citizenship?\\
\midrule
\textbf{Pre-edited Answer $t_5$ and Edited Answer $t'_{3}$} \\
$(t_5)$ Washington, D.C. \\
$(t'_{3})$ Birmingham \\
\midrule
\textbf{Pre-edited Fact Chain} $G_q$ \\
$(\delta_1)$ (\colorbox{LightSkyBlue}{Misery}, author, \colorbox{LightPink}{Stephen King}) \\
$(\delta_4)$ (\colorbox{LightPink}{Stephen King},  citizen of, \colorbox{LemonChiffon}{United States}) \\
$(\delta_5)$ (\colorbox{LemonChiffon}{United States}, capital, \colorbox{LightBlue}{Washington, D.C.}) \\
\midrule
\textbf{Edited Fact Chain} $G^*_q$ \\
$(\delta_1')$ (\colorbox{LightSkyBlue}{Misery}, author, \colorbox{LightSalmon}{Richard Dawkins}) \,\,\, \color{gray}{\#edited fact} \\
$(\delta_2)$ (\colorbox{LightSalmon}{Richard Dawkins}, citizen of, \colorbox{LightGreen}{United Kingdom}) \,\,\, \color{gray}{\#unedited fact}\\
$(\delta_3')$ (\colorbox{LightGreen}{United Kingdom}, capital,\colorbox{Lavender}{Birmingham}) \,\,\, \color{gray}{\#edited fact}\\
\bottomrule
\end{tabularx}
\end{table}
\begin{problem}[Retrieval-Augmented Editing]
Given an edited fact bank $ \Delta = \{\delta_1', \delta_2',...,\delta_N'\}$ with $N$ instances and a multi-hop question $q$ whose answer requires model editing,
we want to retrieve its corresponding edited facts $\Delta_q = \{\delta_i', \delta_j',...,\delta_k'\}$. 
The goal is to ensure that all the edited facts necessary for answering $q$ are retrieved, i.e., $\Delta_q\subseteq G^*_q$ and $\Delta\backslash\Delta_q \not\subset G^*_q$.
Then, these facts are used to refine the target model $f_{\theta}$ for editing.
\end{problem}

\section{Methodology}
\begin{figure*}
    \centering
    \includegraphics[width=0.99\linewidth]{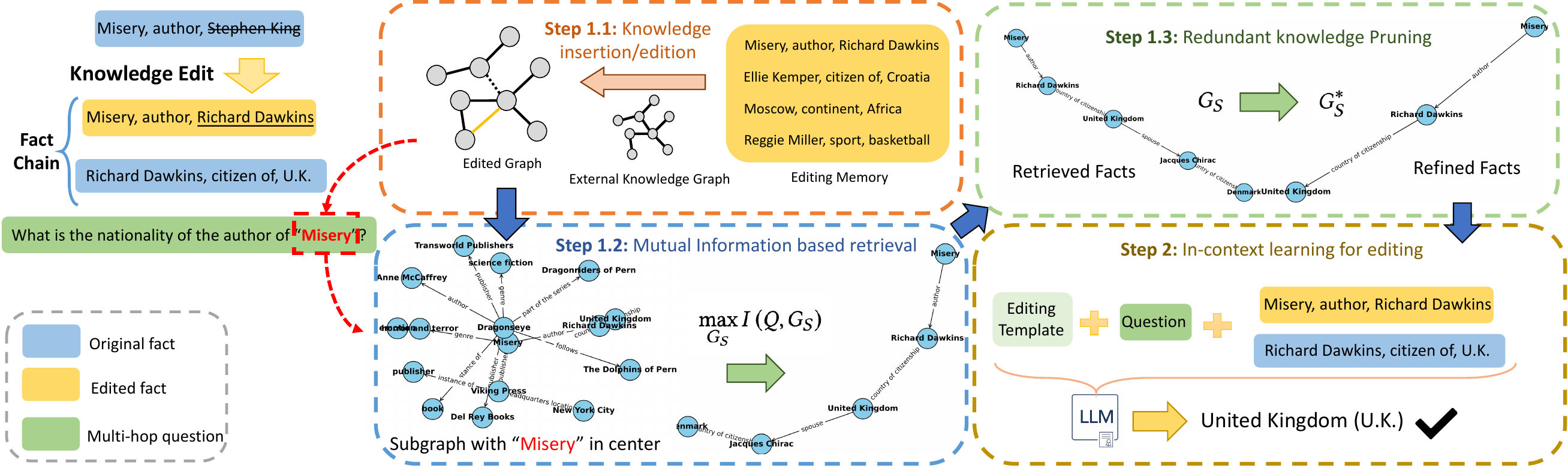}
    \vspace{-0.3cm}
    \caption{The overall framework of our retrieval-augmented in-context model editing method.}
    \label{fig:figure2}
    \vspace{-8pt}
\end{figure*}
Our Retrieval-Augmented Editing (RAE) framework, as shown in Figure~\ref{fig:figure2}, contains two key steps: (1) retrieving edited facts relevant to the question, and (2) editing the language model using these retrieved facts via in-context learning. We will first discuss step~(2) with the motivation of our design in the following section. The details of step~(1) are in Sections~\ref{retrieval} and~\ref{Redundant_Facts_Pruning}.

\subsection{Retrieval-Augmented Editing}
\label{ice}

A na\"ive edit approach uses similarity-based search to retrieve edited facts similar to target question $q$~\cite{wang2023retrieval,han2023improving,zhong2023mquake}. These facts are then integrated into a prompt template for editing via in-context learning:
$
f_{\theta}'(q) = f_{\theta}(T_{e}(q, \{\delta_1', \delta_2',... \delta_K'\})),
$ where $T_e$ is the editing template. For example, $T_e(\cdot)$ can be made as \textit{"Given fact: $\{\delta'\}$, $\{q\}$ ?"}. The Top-$K$ nearest edited facts to question $q$ in the embedding space are denoted as $ \{\delta_1', \delta_2',... \delta_K'\} = {\text{Top-}K}_{\delta \in \Delta} \, \text{sim}(g_{z}(\delta), g_{z}(q))$, where $\text{sim}(\cdot)$ denotes the similarity function and $g_{z}$ is an embedding model.
However, the edited facts $\Delta_q$ needed to answer $q$ are difficult to retrieve by this approach since they usually contain entities different from $q$, which will result in a low similarity score in a large bank $\Delta$ (e.g., in Table~\ref{table1}, \colorbox{LightGreen}{\textit{United Kingdom}} in $\delta_3'$, but not in $q$).

To address this problem, we propose \textbf{edited fact chain extraction} to obtain $G^*_q$. Inherently,
each $G^*_q$ forms a connected knowledge graph (KG)~\cite{zhong2023mquake}. Such KGs can be retrieved by iteratively traversing links from one entity to another.
Take $G^*_q = \{\delta'_1, \delta_2,\delta'_3\}$ in Table~\ref{table1} as an example. It is composed of two edited facts $\delta'_1, \delta'_3$ and one unedited fact $\delta_2$. We can observe that the question entity: (\colorbox{LightSkyBlue}{\textit{Misery}}) is the head entity $h_1$ in $\delta_1' = \{h_1, t_1, t_1'\}$, and the edited tail entity $t'_1$: (\colorbox{LightSalmon}{\textit{Richard Dawkins}}) is also the head entity $h_2$ in the next fact $\delta_2 = \{h_2, r_2, t_2\}$. Moreover, for each subsequent fact in the chain, its head entity is always the tail entity of the previous fact.
By effectively retrieving the KG that represents the fact chain $G^*_q$, we are able to capture all the edited factual triplets $\Delta_q=\{\delta_i', \delta_j',... \delta_k'\}$. In light of this, we define our retrieval-augmented editing as:
\begin{equation}
    f_{\theta}'(q) = f_{\theta}(T_{e}(q, G^*_q)),
\end{equation}
where we give an example of such editing in Figure~\ref{fig:figure2}. 
In the next section, we will introduce the detailed strategy of retrieving $G^*_q$, where we first propose a mutual information-based retrieval strategy to extract facts needed to answer the target question (Section~\ref{retrieval}). Then, we propose a pruning method to delete irrelevant facts from the initial retrieval result (Section~\ref{Redundant_Facts_Pruning}).

\subsection{Edited Facts Retrieval via Maximizing MI}
\label{retrieval}
We first construct a knowledge graph that connects different facts. Then, we introduce our proposed retrieval objective of extracting relevant subgraphs given input questions.

\subsubsection{External Knowledge Graph for Subgraph Retrieval}
According to our previous discussion, we aim to retrieve the fact chain $G^*_q$ for model editing. Additionally, it is worth noting that $G^*_q$ consists of both edited and unedited facts. However, the unedited facts are not included in our edited fact bank $\Delta$ by default.  To effectively incorporate both types of facts into our retrieval process, we propose integrating all edited facts into an external knowledge graph $\mathcal{G}$. By selecting a comprehensive KG such as WikiData~\cite{vrandevcic2014wikidata}, the new graph $\mathcal{G}^*$ will encompass both unedited and edited facts. It complements our edited fact bank $\Delta$ and connects different entities. Besides, the external knowledge graph provides extra factual knowledge that can enhance language to output correct answers.

Specifically, given the edits $\Delta = \{\delta_1', ..., \delta_n'\}$ and an external $\mathcal{G}$, we consider two types of operations to combine them. 
\textbf{(1) Modifying existing facts:} If the original fact appears in $\mathcal{G}$, i.e., $(h,r,t) \in \mathcal{G}$, we will modify the KG according to the edits, so $\mathcal{G}^* = (h,r,t') \cup \mathcal{G}\setminus (h,r,t)$. 
\textbf{(2) Adding new facts:} If the original fact does not appear in $\mathcal{G}$, i.e., $(h,r,t) \notin \mathcal{G}$, then we append the modified fact to the KG, so $\mathcal{G}^* = (h,r,t') \cup \mathcal{G}$. 
Next, given a question $q$, we retrieve a subgraph $G_S$ from $\mathcal{G}^*$, so that $G_S \subset \mathcal{G}^*$. Our goal is to ensure that $G_S$ contains fact chains of $q$, i.e., $G^*_q \subseteq G_S$.

\subsubsection{Mutual Information based Retrieval Objective}
For effective editing, the retrieved subgraph $G_S$ must share relevant information with the question. 
Therefore, we define the objective of subgraph retrieval as maximizing the mutual information (MI) between the subgraph and a set of questions $Q$ whose answers require editing.
The objective is formalized as below, where the theoretical justification is provided in Section~\ref{TJ}:
\begin{equation}
    \max_{G_{S}} I\left(Q; G_{S}\right)=H(Q)-H\left(Q \mid G=G_{S}\right).
    \label{retrieval_objective}
\end{equation}
Given a fixed question set $Q$, its Shannon entropy $H(Q)$ is constant. Therefore, maximizing the mutual information $I\left(Q; G_{S}\right)$ is equivalent to minimizing the conditional entropy $H\left(Q \mid G=G_{S}\right)$. Thus, we optimize the following objective:
\begin{align}
    &\max_{G_{S}} I\left(Q; G_{S}\right)= \min_{G_{S}} H\left(Q \mid G=G_{S}\right)\\
    =&  \max_{G_{S}}  \sum_{q \in Q} p(q | G = G_S )\log_2 p(q | G = G_S ).
\end{align}
In practice, quantifying $p(q | G = G_S )$ is challenging due to its computational complexity. This is because there are numerous subgraph candidates $G_{S}$ within the entire knowledge graph, making it prohibitively expensive to exhaustively search for the optimal one. To circumvent this issue, we first replace the intractable term $p(q | G = G_S )$ with $\frac{p(q, G = G_S)}{p(G = G_S)}$. Then, suppose we consider one question each time, where $Q = {q}$, the objective is reformulated as:
\begin{equation}
    \label{MI}
    \max_{G_{S}} \frac{p(q, G = G_S)}{p(G = G_S)}\log_2 \frac{p(q, G = G_S)}{p(G = G_S)}.
\end{equation}
In the following, we will discuss how to estimate probability $p(q, G = G_S)$ and $p(G = G_S)$ efficiently.

\subsubsection{Probabilities Estimation}
\label{prob_esti}
We propose to compute probabilities by leveraging the \textbf{next-word prediction capability of LLMs}.
Given that the fact chain forms a tail-to-head connected knowledge graph, our extracted subgraph $G_S$ can be represented as $G_S = (h_1, r_1, t_1,...,h_n, r_n, t_n)$, where $h_i$ and $t_i$ are nodes, $r_i$ is the edge, and $n$ is the number of retrieved triplets.
Thus, we can estimate $\frac{p(q, G = G_S)}{p(G = G_S)}$ as:
\begin{equation}
    \label{term_decomp}
    \frac{p(q, G = G_S)}{p(G = G_S)} =  \frac{p(r_1, t_1, h_2, r_2, t_2,...,h_n, r_n, t_n|q, h_1)} {p (r_1, t_1, h_2, r_2, t_2,...,h_n, r_n, t_n|h_1) }\cdot \frac{p(q, h_1)}{p(h_1)} .
\end{equation}
Specifically, for the term $p(r_1, t_1, h_2, r_2, t_2,... | q, h_1)$, we can further decompose it into following form:
\begin{equation}
\label{rel}
\begin{split}
    & p(r_1, t_1, h_2, r_2, t_2,...,h_n, r_n, t_n | q, h_1)\\ 
    =\, &p ( t_1, h_2, r_2, t_2,...,h_n, r_n, t_n |q, h_1, r_1)\cdot p (r_1 | q, h_1).   
\end{split} 
\end{equation}
This decomposition allows us to initially focus on estimating the $p (r_1 | q, h_1)$. Specifically, the head entity $h_1$ is determined if $q$ is given, since we assume $h_1$ is mentioned in question $q$. Candidate relations for $r_1$ can also be selected from the edited KG.
Practically, we can estimate the probability $p (r_1 |q, h_1)$ for each candidate relation using an auto-regressive language model $f_{\phi}$~\cite{radford2019language, wu2023towards}: 
\begin{equation}
    \begin{split}
    &p( r_1 | q, h_1) \approx \\
    &\prod_{i = 1}^{|r_1|} f_{\phi}(w_{r_1}^{(i)}|w_{q}^{(1)},...,w_{q}^{(|q|)}, w_{h_1}^{(1)},...,w_{h_1}^{(|h_1|)}, w_{r_1}^{(1)}, ...,w_{r_1}^{(i-1)}),
\end{split}
\vspace{-0.6cm}
\end{equation}
where $f_{\phi}$ is the predicted word probability, and $w_q, w_{h_1}, w_{r_1}$ denote the words in question $q$, head entity $h_1$, and relation $r_1$, respectively.
We can employ open-source LLMs like GPT-2~\cite{radford2019language} for this estimation. Please note that, the model $f_\theta$ being edited does not need to be the same model used for probability estimation, making our method applicable even for editing proprietary LLMs.
With a specific input context $\{q, h_1\}$, the language model will assign different probabilities to each relation based on its contextual understanding and reasoning ability.

Then, $p ( t_1, h_2, r_2, t_2,...|q, h_1, r_1)$ can be further decompose into $p(h_2, r_2, t_2,...|q, h_1, r_1, t_1)\cdot p(t_1|q, h_1, r_1)$. In our case, we assume $p(t_1|q, h_1, r_1) = 1$, since one relation usually only corresponds to one tail entity. When there are multiple tail entities, we find the assumption still works well empirically.
So, $p(h_2, r_2, t_2,...|q, h_1, r_1, t_1)$ can be decomposed into $p(r_2, t_2,...|q, h_1, r_1, t_1, h_2) \cdot p(h_2|q, h_1, r_1, t_1)$. 
Additionally, since the tail entity in one fact becomes the head entity in the subsequent fact, we can also have $p(h_2|q, h_1, r_1, t_1) = 1$. 
Thus, we can have $p ( t_1, h_2, r_2, t_2,...|q, h_1, r_1) = p(r_2, t_2,...|q, h_1, r_1, t_1, h_2)$. This is a nice property that helps us iteratively decompose this intractable probability term.
By iteratively applying the aforementioned step for $n$ times, we can compute the conditional probability of all subgraphs within an $n$-hop distance from the question entity. The final estimation can be expressed as:
\begin{equation}
\begin{split}
     &p(r_1, t_1, h_2, r_2, t_2,...,h_n, r_n, t_n | q, h_1)  \\
     =\,\, &p ( r_n | q, h_1, r_1, t_1,...,h_{n-1}, r_{n-1}, t_{n-1}, h_{n} ) \cdot\\
     &p ( r_{n-1} |q, h_1, r_1, t_1,...,h_{n-2}, r_{n-2}, t_{n-2}, h_{n-1} )  \cdot \\
     &... \cdot\\
     &p(r_2 | q, h_1, r_1, t_1, h_2)\cdot p (r_1 |q, h_1).
\end{split}
\end{equation} 
Till now, we have decomposed $p(r_1, t_1, h_2,... | q, h_1)$ in Equation~\eqref{term_decomp} into the product of conditional probabilities of predicting different relations within the $n$-hop subgraph. This nice property ensures the selection of the subgraph will only be determined by relation probability, which is free from the interference of any potential edited tail entity.
Similarly, we can decompose the denominator term $p(r_1, t_1, h_2, ...| h_1)$ into:
\begin{equation}
\label{final_form}
\begin{split}
    & p(r_1, t_1, h_2, r_2, t_2,...,h_n, r_n, t_n | h_1)\\ 
    = \, &p ( r_n| h_1, r_1, t_1,...,h_{n-1}, r_{n-1}, t_{n-1}, h_{n} ) \cdot\\
     &p ( r_{n-1}| h_1, r_1, t_1,...,h_{n-2}, r_{n-2}, t_{n-2}, h_{n-1} )\cdot ...\cdot p (r_1 | h_1).
\end{split}  
\end{equation}
Then, for the last term $p(q, h_1)/p(h_1)$ in Equation~\eqref{term_decomp}, based on Bayes' theorem, we can transform it into $p(q, h_1)/p(h_1) = p(q|h_1)$, which is a constant value given a specific question $q$. We can also apply model $f_\phi$ to estimate this conditional probability.
Now, since we are able to estimate every term in Equation~\eqref{MI} and~\eqref{term_decomp}, we can effectively identify the subgraph that yields the maximum Mutual Information. 
Additionally, we utilize beam search~\cite{carnegie1977speech} to expedite the computational process, eliminating the necessity for exhaustively traversing all connected nodes.
Further details on our implementations are provided in the Appendix~\ref{exp_detail}.
In this work, we treat $n$ as a hyperparameter since the number of hops required to answer a question is unknown in advance. To ensure thorough exploration, $n$ is assigned a large value. However, this approach will also introduce irrelevant information in the retrieved subgraph $G_S$, which can potentially mislead the language model to hallucinate and generate undesired answers~\cite{li2024dawn, li2023helma}. In the next section, we will discuss how to mitigate this problem.

\begin{figure}[t]
    \centering
    \begin{subfigure}[b]{0.5\textwidth}
    \begin{tikzpicture}
        \hspace{-0.4cm}
        \begin{axis}[
            ymin=0.22,
            ymax=1.0,
            ybar,
            enlargelimits=0.26,
            symbolic x coords={2-hops, 3-hops, 4-hops},
            xtick={2-hops, 3-hops, 4-hops},
            ytick={0.5,1.0},
            ylabel style={align=center, font=\footnotesize}, ylabel=Normalized Model\\Edit Entropy,
            y tick label style={font=\small},
            x tick label style={font=\small},
            width=0.98\textwidth,
            height=3.5cm,
            bar shift=0cm,
            bar width=0.3cm,
            xtick style={draw=none},
            x tick label style={anchor=center, xshift=0.cm, yshift=-0.1cm},
            x=2.5cm,
            legend style={at={(0.5,1.33)},
                  below,
                  draw=none,
                  legend columns=-1,
                  font=\footnotesize,
                  nodes={text width=width("1st Prefix subset")}
                  /tikz/every even column/.append style={column sep=1em}
                  }
        ]
        \addplot+[bar shift=-0.75cm] coordinates {(2-hops,0.9147) (3-hops,0.9322) (4-hops,0.9453)};
        \addplot+[bar shift=-0.45cm] coordinates {(2-hops,0.0172)  (3-hops,0.6918) (4-hops,0.8153)};
        \addplot+[bar shift=-0.15cm] coordinates {(2-hops,0.2464) (3-hops,0.0175) (4-hops,0.4406)};
        \addplot+[bar shift=0.15cm] coordinates {(2-hops,0.4048) (3-hops,0.1777) (4-hops,0.0099)};
        \addplot+[bar shift=0.45cm] coordinates {(2-hops,0.4889) (3-hops,0.2863) (4-hops,0.1556)};
        \addplot+[draw=applegreen, fill=applegreen, bar shift=0.75cm] coordinates {(2-hops,0.5817) (3-hops, 0.3792) (4-hops,0.2661)};
        \legend{Subset 1\,\,, Subset 2\,\,, Subset 3\,\,, Subset 4\,\,, Subset 5\,\,, Subset 6}
        \end{axis}
    \end{tikzpicture}
     \vspace{-0.5cm}
    \caption{Fact chain $G^*_q$ with redundant knowledge.}
    \label{fig_fact_subset}
    \end{subfigure}
    \vspace{-10pt}
    \newline
     \begin{subfigure}[b]{0.5\textwidth}
   \begin{tikzpicture}
        \hspace{-0.5cm}
        \begin{axis}[
            ymin=0.22,
            ymax=1.0,
            ybar,
            enlargelimits=0.26,
            symbolic x coords={2-hops, 3-hops, 4-hops},
            xtick={2-hops, 3-hops, 4-hops},
            ytick={0.5,1.0},
            ylabel style={align=center, font=\footnotesize}, ylabel=Normalized Model\\Edit Entropy,
            y tick label style={font=\small},
            x tick label style={font=\small},
            width=0.85\textwidth,
            height=3.5cm,
            bar shift=0cm,
            bar width=0.3cm,
            xtick style={draw=none},
            x tick label style={anchor=center, xshift=0.cm, yshift=-0.1cm},
            x=2.5cm, 
        ]
        \addplot+[bar shift=-0.75cm] coordinates {(2-hops,0.5482) (3-hops,0.6163) (4-hops,0.6120)};
        \addplot+[bar shift=-0.45cm] coordinates {(2-hops,0.5143)  (3-hops, 0.4929) (4-hops,0.5561)};
        \addplot+[bar shift=-0.15cm] coordinates {(2-hops,0.5104) (3-hops,0.5166) (4-hops,0.4784)};
        \addplot+[bar shift=0.15cm] coordinates {(2-hops,0.4970) (3-hops,0.4963) (4-hops,0.4825)};
        \addplot+[bar shift=0.45cm] coordinates {(2-hops,0.4935) (3-hops,0.5390) (4-hops,0.5178)};
        \addplot+[fill=applegreen, draw=applegreen, bar shift=0.75cm] coordinates {(2-hops,0.5141) (3-hops,0.5772) (4-hops,0.5319)};
        \end{axis}
    \end{tikzpicture}
    \vspace{-0.7cm}
    \caption{Random fact chain without useful knowledge.}
    \end{subfigure}
\vspace{-0.7cm}
\caption{Distribution of normalized model editing entropy with different fact subsets as input. A lower normalized entropy indicates that the model is more confident in answering the question with the given facts.  "Subset 1" includes the first fact $\{\delta_1\}$, "Subset 2" includes the first two facts $\{\delta_1, \delta_2\}$, and so on. Figure~\ref{fig_fact_subset} shows that the entropy is significantly lower if the subset contains exactly the entire fact chain of the question (e.g., Subset 2 has low entropy for 2-hop questions).} 
\label{fig:editing_entropy}
\end{figure}
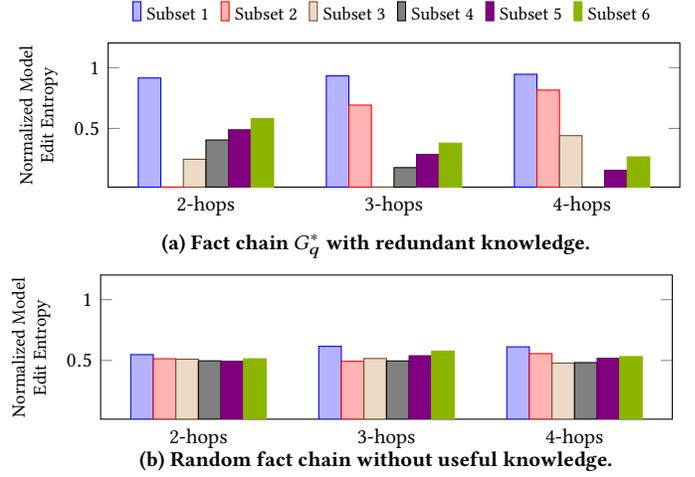

\subsection{Uncertainty-based Redundant Fact Pruning}
\label{Redundant_Facts_Pruning}

This section introduces a pruning method, which utilizes model output uncertainty, to eliminate redundant facts from $G_S$.

\subsubsection{Editing Uncertainty}
We define editing uncertainty as the uncertainty of the output generated by large language models. Formally, the output uncertainty is quantified by Shannon entropy:
\begin{equation}
    H(Y | X = x) = -\sum_{y} p(y | x) \log_2 p(y|x),
\end{equation}
where $y$ represents each possible answer generated by the language model, and $x = \{q, G_S\}$ is the model input composed of the question $q$ and facts $G_S$.  
A higher entropy value $H(Y | X = x )$ means less confidence in the answer, reflecting greater uncertainty. In contrast, a lower entropy value indicates higher confidence and less uncertainty.  
Ideally, if input facts $G_S$ are exactly the edited question fact chain $G^*_q$, i.e., $G_S = G^*_q$, then the model output should exhibit maximum confidence with minimal entropy, since $G^*_q$ contains the precise knowledge to answer question $q$. In the next part, we conduct empirical experiments to verify this assumption.

In our experiments, we choose GPT-J (6B)~\cite{gpt-j} as the base language model. We select 1000 instances for each of the 2, 3 and 4-hop questions from the MQUAKE-CF dataset~\cite{zhong2023mquake} for testing. The MQUAKE-CF dataset comprises multi-hop questions that are based on real-world facts, where the edited facts are counterfactual, meaning they do not exist in actual real-world scenarios. An example of such a question with an edit is provided in Table~\ref{table1}.

Our experiment seeks to identify the fact set $G'_S$ that, when used as model input, yields the lowest output entropy (i.e., minimal editing uncertainty). Our first step is to construct different fact set candidates.
We begin with the first fact $\delta_1$ in the fact chain $G^*_q$ as our initial fact set $G'_S$. Then, we add each subsequent fact from the chain until the $G'_S$ encompasses the entire fact chain $G^*_q$. After that, we insert unrelated facts $\hat{\delta}$ into the set. In this experiment, the process is repeated until $G'_S$ contains six elements, where we build a prefix set $\bar{G}_q$ with all the six subsets $G'_S$. For example, for a $4$-hop question, we have $\bar{G}_q = \{\{\delta_1\}, \{\delta_1, \delta_2\}, \{\delta_1, \delta_2, \delta_3\},...,
\{\delta_1,\delta_2, \delta_3, \delta_4, \hat{\delta}_5, \hat{\delta}_{6}\}\}$, where $\delta_1,\delta_2, \delta_3, \delta_4 \in G^*_q$ and $\hat{\delta}_5, \hat{\delta}_{6}$ are two irrelevant facts.
Finally, we conduct in-context editing using each subset $G'_S$ from $\bar{G}_q$ with editing template $T_e$:"\textit{Given fact: $\{G'_S\}$, $\{q\}$?}". 
In our experiment,
we consider model output $y$ to be each of the next predicted word. We report the entropy over all the words in the vocabulary as the editing uncertainty.

The editing uncertainty with different subsets is listed in Figure~\ref{fig:editing_entropy}a. For comparison, we also report the editing uncertainty with random facts selected from Wikidata, as shown in Figure~\ref{fig:editing_entropy}b. Our observations reveal a phenomenon: the language model produces answers with much lower entropy when $G'_S$ is equal to the ground-truth fact chain $G^*_q$. If $G'_S$ presents redundant facts or insufficient facts, the entropy will increase. Meanwhile, if the LLM is fed with random facts, the entropy level also remains consistently high. \textbf{This observation justifies the use of entropy as the indicator of whether $G'_S$ contains the correct facts for LLM inference.}

\subsubsection{Knowledge Pruning with Editing Uncertainty}
Since incorporating the most relevant facts will result in the lowest entropy, we propose to utilize this finding for knowledge pruning. Specifically, we first follow Section~\ref{retrieval} to retrieve a knowledge graph $G_S$ containing $n$ triplets for question $q$: $G_S = \{\delta_1, \delta_2, ..., \delta_n\}$, where $n$ is a sufficiently large number, and $G_S$ could contain redundant knowledge. 
Then, to remove redundant knowledge, we first build the prefix sets $\bar{G}_q$ for target question $q$ based on the retrieved graph $G_S$. Then, we can obtain the pruned fact set $G^*_S$ using the objective: 
\begin{equation}
    G^*_S = \argmin_{G'_S \in \bar{G}_q} -\sum_{y} p(y | T_e(q, G'_S)) \log_2(p(y|T_e(q, G'_S))).
\end{equation}
Finally, we can apply $G^*_S$ as our retrieved fact chain for the in-context learning introduced in Section~\ref{ice} to conduct editing.

\subsection{Theoretical Justification}
\label{TJ}
In this subsection, we theoretically justify that the facts collected by our retrieval objective Eq.~\eqref{retrieval_objective} are effective in performing model editing with in-context learning. To begin with, we discuss what kinds of input can effectively activate in-context learning. Then, we explore how to build such effective input for model editing. 

Our proposed editing method relies on the in-context learning ability of LLMs. In the following, we provide an analysis of how in-context learning can be effectively triggered.
Theoretically, the text generation process of a language model can be understood as a Hidden Markov Model~\cite{baum1966statistical, xie2021explanation}. The model initially selects a concept $\theta_c \in \Theta$ from a set of underlying concepts denoted as $\Theta$, and then samples a sequence of words based on the chosen concept.
Based on that, the in-context learning can be written as $ p(y|S, x) = \int_{\theta_c\in\Theta}p(y|S, x,\theta_c)p(\theta_c|S, x)d\theta_c$, where $S$ denotes in-context prompt and $x$ denotes query.
Existing research has theoretically proven that the condition to activate in-context learning is when there is a shared latent concept $\theta_c$ between prompt text $S$ and the input query $x$. More discussions can be found in~\citep{xie2021explanation}. 

Motivated by the above analysis, as in-context prompt $S$ is the edited knowledge in our design, we seek to include the edited knowledge that shares the same latent concept $\theta_c$ as question $q$.
Ideally, this will activate in-context learning for effective model editing. 
Formally, we can define such knowledge graph as 
\begin{equation}
    G_S = \argmax_{G \in \mathcal{G}}I(G; \theta_c),
\end{equation}
where $\theta_c$ is the latent concept used to generate question $q$, and we use mutual information $I(G; \theta_c)$ to quantify the share information. However, this is a non-trivial task since concept $\theta_c$ is an intractable hidden variable. To address this issue, we propose obtaining the target knowledge graph that maximizes the lower bound of such an objective. Specifically, we can have the following theorem:
\begin{theorem} Given retrieved graph $G_S \in \mathcal{G}$, the latent concept $\theta_c$, and the question $q$ sampled conditioned on concept $\theta_c$, there exists a mutual information inequality:
\begin{equation}
    I(G_S; \theta_c) \geq I(G_S; q).
\end{equation}
\label{theorem_eq}
\vspace{-0.5cm}
\end{theorem}
Theorem 1 shows that we can maximize the mutual information between the selected knowledge graph $G_S$ and question concepts $\theta_c$ by maximizing the mutual information between the selected graph $G_S$ and the question $q$ itself. In this way, the in-context learning ability of LLMs would be effectively triggered.
When we apply such knowledge as the prompt, we can effectively conduct the in-context editing. The proof of Theorem~\ref{theorem_eq} is in Appendix~\ref{proof}.


\section{Experiments}

We conduct experiments to answer the following questions. \textbf{Q1}: How effective is RAE in editing LLM output? \textbf{Q2}: How does our retrieval strategy perform compared to other retrieval methods? \textbf{Q3}: Does our proposed pruning technique remove redundant facts from the retrieved facts? \textbf{Q4}: Does RAE work for propriety LLMs?

\subsection{Experiment Settings}

\subsubsection{Language Models}

We evaluate RAE across various kinds of language models in different sizes and families, including GPT-2 (1.5B) ~\cite{radford2019language}, GPT-J (6B)~\cite{gpt-j}, Falcon (7B)~\cite{almazrouei2023falcon}, Vicuna (7B)~\cite{vicuna2023}, and Llama2-chat (7B)~\cite{touvron2023llama2}. Among them, GPT-2, GPT-J, and Falcon are pre-trained language models without instruction tuning~\cite{ouyang2022training, chung2022scaling}, while Vicuna is an instruction-tuned variation of Llama1~\cite{touvron2023llama} and Llama2-chat is the instruction-tuned version of Llama2. Instruction-tuned models (Vicuna and Llama2-chat) are expected to better follow the instructions in the prompt compared to native pre-trained models (GPT-2, GPT-J, and Falcon). We include both kinds of models to verify the effectiveness of the proposed methods.

\subsubsection{Editing Baselines}
For comparison, we consider three kinds of model editing methods: (1) Model weight updating methods: Fine-tuning~\cite{zhu2020modifying} edits the model weights by language modeling the edited knowledge. ROME~\cite{meng2022locating} and MEMIT~\cite{meng2022mass} focus on identifying and updating particular neurons associated with the knowledge that needs editing. (2) Auxiliary models methods: SEARC~\cite{mitchell2022memory} trains an extra language model to store updated knowledge, and it switches to the auxiliary model when answering questions relevant to the edited facts. 
(3) RAG-based methods: Mello~\cite{zhong2023mquake} and DeepEdit~\cite{wang2024deepedit} represent cutting-edge editing methods for multi-hop questions, employing multi-round conversations to edit model outputs. Additionally, the Subgraph Retriever (SR)~\cite{zhang2022subgraph} introduces an advanced knowledge retrieval approach for multi-hop question-answering tasks. We adapt their retrieval method as a baseline. 

\begin{table*}[t]
\renewcommand\arraystretch{1}
\caption{Edited accuracy (\%) on multi-hop question editing datasets.}
\vspace{-0.4cm}
\label{cf}
\begin{tabular}{cl|c|c|c|c|c|c|c|c}
\toprule
& &   \multicolumn{7}{c}{Editing Methods}\\ \hline
\multicolumn{1}{c}{Language Models}&Datasets & Fine Tune& ROME & MEMIT& SEARC& Mello &DeepEdit &\tabincell{c}{Subgraph\\Retriever}  &RAE(ours)\\ \hline
\multirow{3}{*}{\tabincell{c}{GPT-2 (1.5B)}}& M-CF&  3.8& 1.7& 2.3& 4.0& 0.0&0.0 &21.9 &\textbf{62.8}\\
 & M-T&  5.8& 6.4& 1.6& 2.7&0.0 &0.0 &20.3 &\textbf{61.8}\\
 & Popular&6.2   &4.3  &2.9  &1.1  &0.0  &0.0 &26.7 &\textbf{47.1}\\ \hline
 \multirow{3}{*}{\tabincell{c}{GPT-J (6B)}}& M-CF&  7.7& 7.6 & 8.1 & 6.8& 15.3&9.3 &36.2 &\textbf{69.3}\\
 & M-T& 3.1&4.1& 10.6& 2.8& 36.7&19.6 &51.2 &\textbf{63.9}\\ 
 & Popular&6.8   &7.5  &4.4  &1.3  &12.8  &6.6 &45.8 &\textbf{49.6}\\ \hline
 \multirow{3}{*}{\tabincell{c}{Falcon (7B)}}& M-CF&  5.6&  1.7& 2.3& 7.9& 10.7&10.8 &40.1 &\textbf{66.8}\\
 & M-T& 17.2& 7.3& 1.6& 4.5& 51.5&31.7 &56.1 &\textbf{61.6}\\
 & Popular&2.1   &4.0  &1.1  &3.0  &8.1  &9.5 &43.0 &\textbf{50.0}\\\hline 
\multirow{3}{*}{\tabincell{c}{Vicuna (7B)}}& M-CF&  4.8& 8.4 & 7.6 & 7.9& 10.2&11.4  &39.4 &\textbf{67.2}\\
 & M-T&  23.1& 5.0& 1.7& 4.5& 51.7&40.4 &58.6 &\textbf{63.2}\\   
 & Popular&4.0   &3.8  &2.4  &3.0  &7.7  &8.2 &29.5 &\textbf{36.1}\\ \hline
 \multirow{3}{*}{\tabincell{c}{Llama2\\(chat) (7B)}}& M-CF& 5.4& 6.3& 3.8& 7.9& 20.7&11.2 &45.7 &\textbf{69.1}\\
 & M-T& 17.1& 8.7& 1.7& 4.5& 49.4&37.9 &63.1 &\textbf{66.2}\\
 & Popular&5.2   &13.8  &4.9  &3.0  &13.5  &11.1 &41.9 &\textbf{51.4}\\
                                          \bottomrule
\end{tabular}
\vspace{-0.3cm}
\end{table*}

\subsubsection{Implementation Details}
We evaluate our editing method on the MQUAKE-CF and MQUAKE-T datasets from~\cite{zhong2023mquake} and Popular datasets from ~\cite{cohen2023evaluating}. The MQUAKE-CF (M-CF) comprises counterfactual editing instances in 2, 3, and 4-hop questions, totaling 3000 edits. The MQUAKE-T dataset (M-T) features temporal editing examples in 2 and 3-hop questions, with a total of 1868 edits. Additionally, the Popular dataset contains counterfactual editing in 2-hop questions, comprising 274 edits.
Following previous work~\cite{zheng2023can, zhong2023mquake}, we leverage relevant cases from the MQUAKE-CF-9k dataset to craft prompt templates for both baselines and our method.
We evaluated our editing method using the multi-hop edited accuracy metrics from~\cite{zhong2023mquake, wang2024deepedit}. 
The results, which reflect the accuracy when all edits are applied in one batch, are reported in Table~\ref{cf}.


\subsection{Editing Performance Evaluation}
To answer \textbf{Q1}, we assess our model editing method across various language models, compared against different baseline methods.
Our key observations from Table~\ref{cf} are: \textbf{(1)} Our RAE outperforms all others in three datasets across five language models when conducting thousands of edits at the same time.
This superior performance primarily stems from our novel MI-based retrieval objective and an effective pruning strategy. Our design can also seamlessly integrate an external knowledge graph, which effectively links all edited facts, thereby facilitating the multi-hop editing process.
\textbf{(2)} RAG-based methods generally show better performance than other methods. Specifically, Mello and DeepEdit demonstrate good performance on the M-T dataset with models larger than 6B. However, they underperform on the M-CF and popular datasets and fail with GPT-2, likely due to GPT-2's inability to follow their complex prompts, resulting in no output. Additionally, the subgraph retriever ranks second in performance. However, its probability-driven retrieval method occasionally fails to fetch relevant facts, leading to reduced effectiveness. 
\textbf{(3)} Other methods generally show lower performance across all language models, which aligns with the findings from~\cite{zhong2023mquake}.

\begin{table}[tp]
\footnotesize
\centering
\caption{Multi-hop facts retrieval precision (\%) comparison.}
\vspace{-0.4cm}
\label{table3}
\begin{tabular}{cl|cc|cc|cc} \toprule
  \multicolumn{8}{c}{MQUAKE-CF}                                                       \\ \hline
 \multicolumn{2}{c|}{Question Type}& \multicolumn{2}{c|}{2-hops} & \multicolumn{2}{c|}{3-hops} & \multicolumn{2}{c}{4-hops} \\ \hline
\multicolumn{1}{c|}{Category} & Retrieval & \multicolumn{1}{c|}{P@1} & \multicolumn{1}{c|}{P@2} & \multicolumn{1}{c|}{P@1} & \multicolumn{1}{c|}{P@3}  & \multicolumn{1}{c|}{P@1} & \multicolumn{1}{c}{P@4}    \\ \hline
\multirow{3}{*}{Embedding} &KG Link           &        52.7&         28.7&         18.2&        3.7&         14.0&                0.0 \\
 &QR   &        62.3&         7.7&         14.7&        0.0&         12.3&                0.0\\ 
& Mello(Llama2)    &        84.3&         80.0&         80.7&       42.3&         83.3&                25.7\\  \hline
\multirow{2}{*}{Probability} &SR(GPT-2)       &        77.7&         50.3&         67.3&        25.3&         65.0&                20.0\\
 &SR(Llama2)        &        78.3&         55.7&         79.7&        37.0&         69.3&                28.7\\ \hline
\multirow{5}{*}{\tabincell{c}{Mutual \\ Information}} &RAE(GPT-2)      &        83.0&         66.3&         77.3&        41.0&         80.3&                43.7\\
 &RAE(GPT-J)             &        83.0&         69.7&         81.3&        53.7&         82.7&            54.0\\
 &RAE(Falcon)&        82.3&         70.7&         72.3&        44.3&         81.7&                47.3\\
 &RAE(Vicuna)&        81.0&         66.7&         79.3&        50.3&         85.0&            50.0\\
 &RAE(Llama2)&        82.7&         69.3&         84.0&        49.3&         82.0&                47.0\\
\bottomrule    
\end{tabular}
\vspace{-0.5cm}
\end{table}

\subsection{Retrieval Performance Evaluation}
To answer \textbf{Q2}, we assess the effectiveness of our mutual information-based retrieval method for multi-hop question-answering tasks. 

\subsubsection{Retrieval Baselines} 
We consider three embedding-based and one probability-based methods as baselines.

(1) Embedding-based methods mostly utilize dense retrieval to fetch relevant corpus from vector databases. Here, we adapt them to fit our multi-hop knowledge graph retrieval task. We employ a cutting-edge retriever, the Contriever~\cite{izacard2021unsupervised}, to encode all edited facts into embeddings, and then cache these embeddings for similarity search. In addition to Mello, we include two representative multi-hop retrieval methods as follows:  
\begin{itemize}[leftmargin=*, topsep=0mm]
\item \textbf{KG Link}~\cite{das2019multi, xiong2019simple} is a straightforward strategy, which identifies the query entity and its linked entities as candidates to find the one that is the most similar to the original question. This process is repeated $K$ times to retrieve the entire fact chain. 
\item \textbf{Question Reform (QR)}~\cite{yadav2020unsupervised, nie2020answering} appends the retrieved entity to the previous query for the next hop retrieval. This design is motivated by simulating a reasoning path where the retrieved fact at each hop is viewed as an intermediate reasoning result. 
\end{itemize}

(2) Probability-based method (i.e., subgraph retriever~\cite{zhang2022subgraph}) retrieves the subgraph $G_S\in \mathcal{G}^*$ that maximizes the conditional probability $p(G_S|q)$, where $q$ is the input multi-hop question.

\subsubsection{Implementation Details}
Following~\cite{schutze2008introduction}, we select 300 cases for each of the 2, 3, and 4-hop questions from the M-CF dataset. For each type, we report the retrieval precision scores in Table~\ref{table3}. Specifically, we use the metric $ Precision@K $, which calculates the proportion of relevant facts within the top $K$ results~\cite{schutze2008introduction}. Here, $ Precision@K = |\{\text{relevant facts}\}|/{K} \times 100\%$, abbreviated as $P@K$. For this study, a $k$-hop question has $k$ facts as its fact chain. A fact is considered relevant if it is part of the fact chain used to answer a multi-hop question.

\subsubsection{Results}
We have the following observations: \textbf{(1)} Our proposed mutual information-based retrieval method demonstrates excellent performance across various LLMs in multi-hop fact extraction. We also find its success with relatively small language models, such as GPT-2, showing strong generalization ability. \textbf{(2)} In contrast, traditional embedding-based methods (KG Link and Question Reform) underperform in this multi-hop fact retrieval challenge. Their limitation mainly lies in failing to comprehend the complex interplay between multiple relations in a question. Thus, it hinders the extraction of target facts from the extensive knowledge base. 
\textbf{(3)} Mello demonstrates the efficacy of decomposing multi-hop questions into single-hop questions for this multi-hop facts retrieval task. Notably, its performance drops significantly with an increasing number of hops, probably because it becomes much more challenging for language models to perform question decomposing.
\textbf{(4)} Probability maximization-based methods outperform traditional embedding methods, while there is still a gap compared to ours, probably because the predicted most probable fact may not be the most necessary one for answering a specific question.

\begin{table}[tp]
\renewcommand\arraystretch{1.0}
\small
\caption{Edited accuracy (\%) with (w) or without (w/o) pruning.}
\label{tbl_pruning}
\vspace{-0.4cm}
\begin{tabular}{l|c|c|c|c|c|c} \toprule
Dataset           & \multicolumn{6}{c}{MQUAKE-CF}          \\ \hline                                             
Type  &Strategy  &GPT-2 &GPT-J &Falcon & Vicuna&  \tabincell{c}{Llama2\\(chat)}  \\ \hline
\multirow{3}{*}{2-hops}&          w/o Pruning&63.0&63.7&          65.2&63.8&          70.1\\ 
                      &          w/ Pruning&73.3&75.5&          74.5&73.5&          75.8\\ 
 &   Gain&16.3\%$\uparrow$&18.5\%$\uparrow$&  14.3\%$\uparrow$&15.2\%$\uparrow$&  8.1\%$\uparrow$\\ \hline
\multirow{3}{*}{3-hops}&          w/o Pruning&43.1&53.8&          55.6&55.0&          60.3\\
 &   w/ Pruning&53.2&65.4&  62.1&62.7&  65.8\\
 & Gain& 23.4\%$\uparrow$& 21.6\%$\uparrow$&11.7\%$\uparrow$&14.0\%$\uparrow$& 9.1\%$\uparrow$\\ \hline
\multirow{3}{*}{4-hops}& w/o Pruning& 49.9& 58.8& 55.2& 61.5& 61.6\\
 & w/ Pruning& 61.9& 66.9& 62.9& 65.5& 65.8\\
 & Gain&24.0\%$\uparrow$&13.8\%$\uparrow$&13.9\%$\uparrow$&6.5\%$\uparrow$&6.8\%$\uparrow$\\     \bottomrule  
\end{tabular}
\vspace{-0.3cm}
\end{table}

\subsection{Ablation Studies on Pruning Strategy}
To answer \textbf{Q3}, we verify that our proposed pruning strategies are beneficial to multi-hop editing tasks. 
To simulate the situation where the retrieved facts contain redundant information, we conduct our experiment by always retrieving 2 additional facts over the facts needed by the original question. That is, given a $k$-hop question, we set the total number of retrieved facts as $n=k+2$. 

Table~\ref{tbl_pruning} reports the edited accuracy of RAE working with or without the pruning strategy, and we draw the following conclusions: 
\textbf{(1)} The proposed pruning technique significantly enhances the performance of model editing, demonstrated by achieving an average accuracy improvement of 14.5\% across various language models. 
\textbf{(2)} We observe a more profound improvement for smaller language models on complex questions, while this benefit to larger language models is relatively less significant. In particular, the performance improvement of GPT-2 on the 4-hop questions reaches 24.0\%, while for Llama-2 it is just 6.8\%. This observation suggests that larger models are more robust to redundant information. 


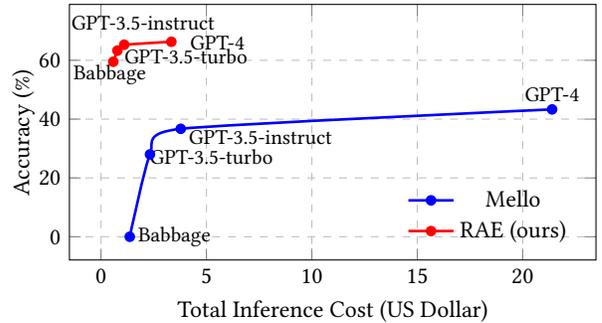
\begin{figure}
    \centering
\begin{tikzpicture}
\hspace{-0.2cm}
    \begin{axis}[
        xlabel={Total Inference Cost (US Dollar)},
        ylabel={Accuracy (\%)},
        ymax=79,
        ylabel style={at={(axis description cs:-0.05,0.5)},anchor=south},
        legend pos=south east,
        grid=major,
        grid style={dashed,gray!30},
        major grid style={lightgray},
        legend style={draw=none, fill=none},
        every axis plot post/.append style={
            mark=*, 
            mark options={solid, mark size=1.5pt},
            line width=1pt
        },
        x tick label style={
            /pgf/number format/fixed,
            /pgf/number format/precision=3
        },
        scale only axis,
        width=7cm,
        height=3.4cm,
    ]
    
    \addplot[color=blue, smooth,tension=0.35] coordinates {
        (1.3648,0.0)   
        (2.324243,28.00)
        (3.78312,36.70) 
        (21.3928,43.30)
    };
    \addlegendentry{Mello}

    \addplot[color=red, smooth,tension=0.1] coordinates {
        (0.583781333,59.5)  
        (0.780253333,63.3)  
        (1.101123333,65.3)  
        (3.336833333,66.3)
    };
    \addlegendentry{RAE (ours)}
    \node[right, font=\small] at (axis cs:1.3648,0.01) {Babbage};
    \node[above, font=\small] at (axis cs:5.24243,22.00) {GPT-3.5-turbo};
    \node[right, font=\small] at (axis cs:3.78312,34.00) {GPT-3.5-instruct};
    \node[above, font=\small] at (axis cs:21.3928,43.30) {GPT-4};
    \node[right, font=\small] at (axis cs:3.836833333,66.0) {GPT-4};
    \node[above, font=\small] at (axis cs:2.0,67.5) {GPT-3.5-instruct};
    \node[right, font=\small] at (axis cs:0.72,61.5) {GPT-3.5-turbo};
    \node[right, font=\small] at (axis cs:-1.7,55) {Babbage};
    \end{axis}
\end{tikzpicture}
\vspace{-10pt}
    \caption{Editing performance and inference cost over different proprietary models.}
    \label{fig:cost and performance}
    \vspace{-0.5cm}
\end{figure}

\subsection{Editing Performance on Proprietary LLMs}
To answer \textbf{Q4}, we apply the proposed RAE to edit proprietary language models, where we can only access the model via APIs, such as ChatGPT~\cite{gpt35}. In such cases, RAE utilizes a different lightweight language model to perform relevant facts retrieval and pruning as introduced in Section~\ref{retrieval} and Section~\ref{Redundant_Facts_Pruning}. Then, the obtained facts will be fed into the proprietary models to perform in-context editing. 
We evaluate our proposed editing method on \textit{GPT-babbage-002}, \textit{GPT-3.5-turbo-0613}, \textit{GPT-3.5-instruct}, and \textit{GPT-4-0613} models~\cite{openai_gpt35}. We use GPT-2 (1.5B) as our retrieval model. We report the edited accuracy and total editing cost of our method for 300 randomly selected cases (MQUAKE-CF) in Figure~\ref{fig:cost and performance}.  
The editing cost includes the total fees of calling APIs and the cost of running GPT-2 for knowledge retrieval on rented GPUs. 
We also report the results of Mello~\cite{zhong2023mquake} for comparison, where only API fees are counted. 

In Figure~\ref{fig:cost and performance}, we first observe that Babbage has 0.0\% edited accuracy by using Mello, showing its ineffectiveness in editing the un-instruction tuned proprietary language model. This is expected since Mello relies on the conversational ability of language models to decompose multi-hop questions. In contrast, RAE is effective in editing all these proprietary models with a remarkably lower cost than Mello. In particular, RAE improves Mello for editing GPT-4 with almost 20\% edited accuracy by only costing around its 15\% budget. 
This highlights the benefit of utilizing the inherent language modeling ability instead of the instruction following ability to perform knowledge retrieval for multi-hop question answering.

\subsection{Performance with Different Batch Sizes}
\begin{figure}[h]
    \vspace{-10pt}
    \begin{subfigure} [t]{0.24\textwidth}
    \includegraphics[width=0.98\textwidth]{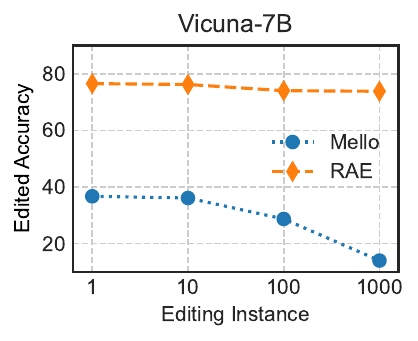}
    \end{subfigure}
    \hspace{-11pt}
    \begin{subfigure} [t]{0.24\textwidth}
    \includegraphics[width=0.98\textwidth]{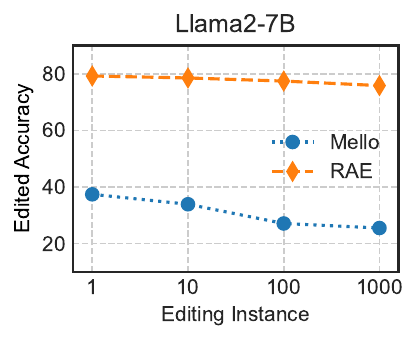}
    \end{subfigure}
    \vspace{-10pt}
    \caption{Edited accuracy with different edit batch sizes.} \label{batch_performance}
    \vspace{-15pt}
\end{figure}

In this section, we evaluate editing performance with different editing batch sizes. Specifically, we set the editing instances in sizes of 1, 10, 100, and 1000 cases for 2-hop questions. We choose Mello~\cite{zhong2023mquake} as our baseline. The result is presented in Figure~\ref{batch_performance}. We can observe that, in both the Vicuna and Llama2 models, RAE's accuracy remains stable across different editing instances, whereas Mello's accuracy significantly declines with increasing instances.

\subsection{Case Study}
\begin{figure}[htp]
    \centering
    \begin{subfigure}[b]{0.49\textwidth}
        \centering
        \includegraphics[width=\textwidth]{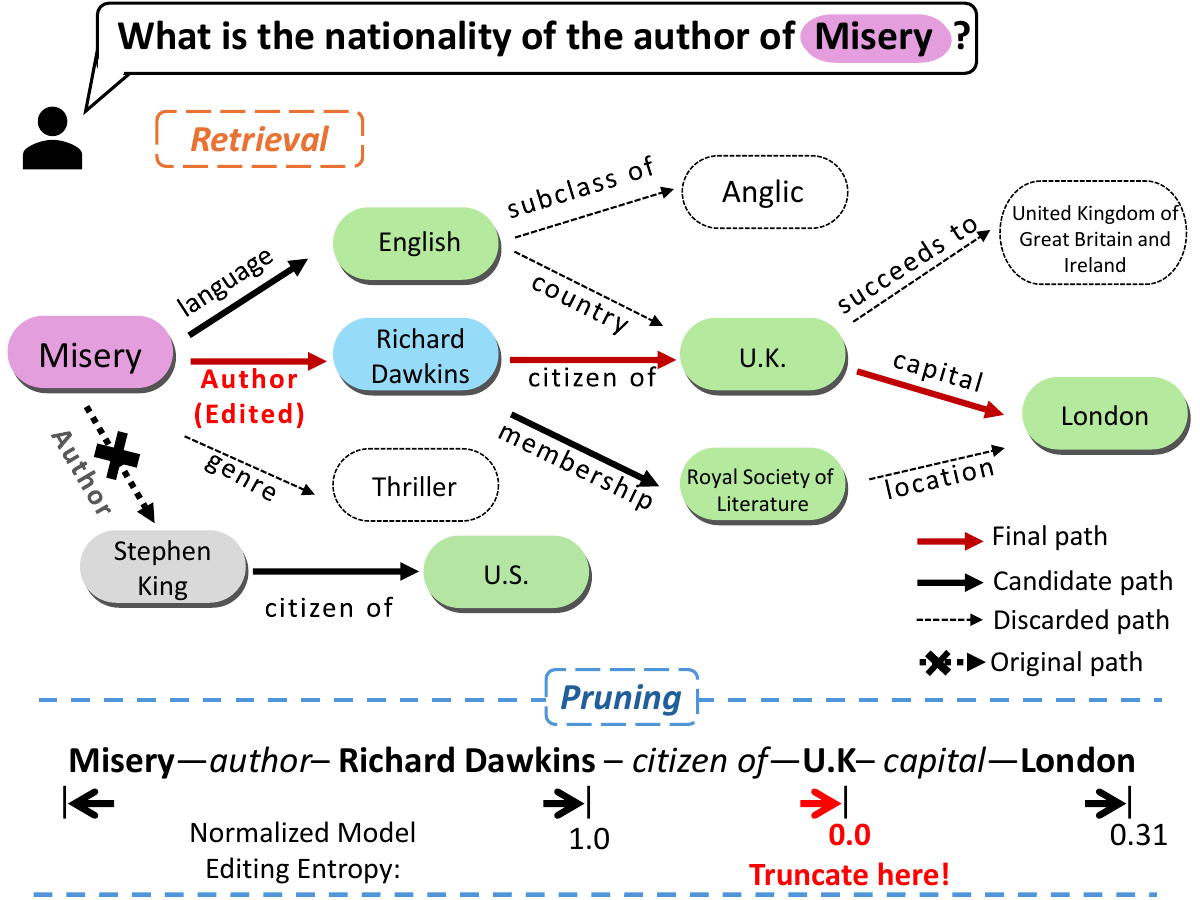}
        \caption{A two-hop question example (CaseID 2).}
        \label{fig:sub1}
    \end{subfigure}
    \vspace{0.2cm}
    \begin{subfigure}[b]{0.49\textwidth}
        \centering
        \includegraphics[width=\textwidth]{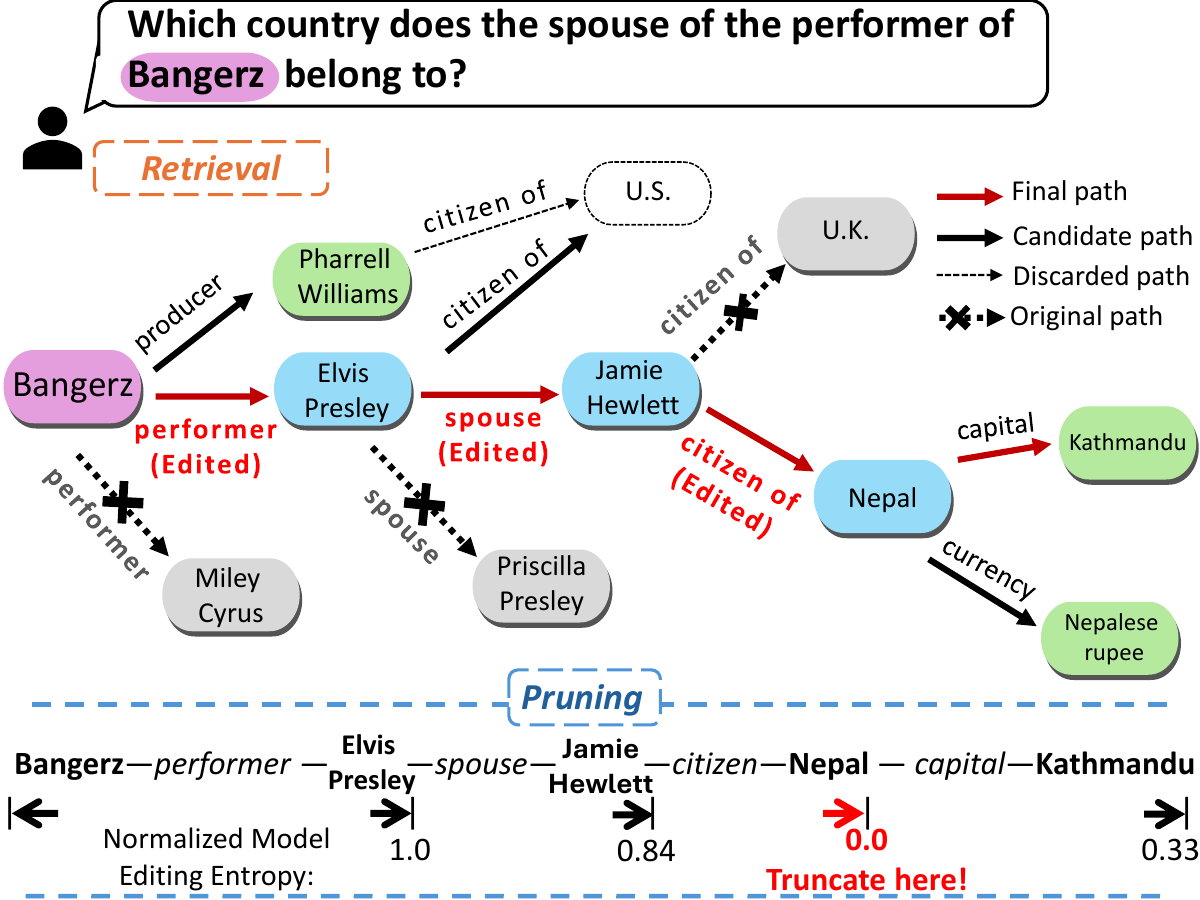}
        \caption{A three-hop question (CaseID 1143).}
        \label{fig:sub2}
    \end{subfigure}
    \vspace{-12pt}
    \caption{Case studies for edited facts retrieval and pruning. The retrieval process involves the beam search, starting from the query entity and navigating through the knowledge graph with two beams. At each entity hop, the two primary candidate edges are highlighted in bold, while others are discarded and marked with dashed lines. The beam search result with the highest MI score is emphasized in red.}
    \label{fig:case_study}
    \vspace{-0.5cm}
\end{figure}

In Figure~\ref{fig:case_study}, we present two cases from the M-CF dataset to demonstrate the retrieval process on a knowledge graph and the pruning process of the retrieved facts. 
For visualizations, the red, black, and dotted lines represent the final, candidate, and discarded paths in the knowledge graph with beam search, reflecting the decision-making process of our retrieval design. 
In Figure~\ref{fig:sub1}, a potential path \textit{Misery--language-->English--country-->U.K.} is discarded even though it can lead to the correct answer (U.K. in this case). This is because it does not share the information that is needed to answer the target question, resulting in a low MI score. In Figure~\ref{fig:sub2}, even though this question is associated with three edited facts, our method can successfully locate all of them, demonstrating robustness over multi-hop questions. In both cases, our pruning strategy successfully truncates the retrieved fact chain to only contain necessary facts relying on the normalized model editing entropy. 

\section{Related Work: Model Editing}


Existing editing methods can be categorized into the following kinds: First, methods that alter model parameters, including fine-tuning~\cite{zhu2020modifying}, locate-then-edit~\cite{meng2022locating, meng2022mass}, and meta-learning~\cite{mitchell2021fast,zhang2024knowledge}, are prone to catastrophic forgetting issue~\cite{dai2021knowledge, zhai2023investigating, cohen2023evaluating}, where the previously encoded knowledge could be lost after editing. They also struggle with using updated language for knowledge reasoning, which can lead to severe hallucinations.~\cite{gekhman2024does, zhong2023mquake, chen2024editingllmsinjectharm}.  While these methods perform well in single-hop editing, they are less effective in multi-hop scenarios.~\cite{zhong2023mquake}. 
Second, methods that depend on training auxiliary models also fall short in these scenarios. The auxiliary models are usually smaller language models, which lack the necessary reasoning capability to infer correct answers~\cite{mitchell2022memory}. 
In contrast, a third category of methods, based on Retrieval-Augmented Generation (RAG), modifies model outputs in a more effective manner~\cite{ zhong2023mquake, yin2024history, wang2024deepedit, gu2024pokemqaprogrammableknowledgeediting}. These methods integrate updated knowledge directly into the model input and edit LLMs through an editing prompt.~\cite{zheng2023can, cohen2023evaluating}. 
RAG-based approaches offer notable benefits as they are resistant to catastrophic forgetting, and allow for on-the-fly editing.~\cite{wang2023retrieval, han2023improving}.
For multi-hop editing, existing RAG-based methods perform well on powerful proprietary LLMs like GPT-3.5, but show significant degradation on less robust LLMs such as Llama2-7b, particularly as the number of editing instances increases. In contrast, our RAE can maintain good editing performance even with relatively small LLMs and large editing batch sizes.

\section{Conclusion}
We propose a novel LLM editing framework for multi-hop QA that employs mutual information maximization for fact retrieval and a self-optimizing technique to prune redundant data. This effectively tackles the integration of real-time knowledge updates in LLMs. Our extensive evaluations confirm the framework's ability to enhance the accuracy of LLM responses, marking significant progress in model editing and dynamic knowledge integration research.

\begin{acks}
The work is, in part, supported by NSF (\#IIS-2223768). The views and conclusions in this paper
are those of the authors and should not be interpreted as representing any funding agencies.
\end{acks}

\appendix
\setcounter{theorem}{0}
\section{Theoretical Justification}
\label{proof}
\begin{theorem} Given retrieved graph $G_S \in \mathcal{G}$, the latent concept $\theta_c$,
and the question $q$ sampled conditioned on concept $\theta_c$, there exists a mutual information inequality: $I(G_S; \theta_c) \geq I(G_S; q).$
\vspace{-0.2cm}
\end{theorem}
\begin{proof}
Consider latent concepts $ \theta_c $ is inferred from a knowledge graph $ G_S $, and a question $ q $ is then generated based on this concept. 
The following Markov chain: $ G_S \rightarrow \theta_c \rightarrow q $ exists, indicating that $q$ is conditionally independent to $G_s$.
According to the chain rule for mutual information, we can have:
\begin{equation}
I(G_S; \theta_c, q) = I(G_S; q) + I(G_S; \theta_c | q)   = I(G_S; \theta_c) + I(G_S; q | \theta_c). 
\end{equation}
Given that the question $ q $ is conditionally independent of the graph $ G_S $, the term $ I(G_S; q | \theta_c) $ equals zero. Therefore, we are left with:
$I(G_S; \theta_c) \geq I(G_S; q).$
Equality holds precisely when $ I(G_S; q | \theta_c) $ is zero, meaning that the graph $ G_S $ provides no additional information about $ q $ once $ \theta_c $ is known.
\end{proof}

\section{Implementation Details}
\label{exp_detail}
In our fact retrieval process, we employ a beam search with a width of 2, meaning we consider only two candidate relations at each hop. For the total number of retrieval hops $n$, we set it to 4, 5, and 6 for 2-hop, 3-hop, and 4-hop questions, respectively. For simplicity, we omit the term $p(q | h_1)$ as it remains constant for each question, and its exclusion does not impact empirical performance. Regarding the SEARC baseline, we pair smaller models with larger auxiliary models: GPT-2-small (124M)~\cite{radford2019language} with GPT-2-xl (1.5B)~\cite{radford2019language}, GPT-2-large (774M)~\cite{radford2019language} with GPT-J (6B)~\cite{gpt-j}, and GPT-Neo (2.7B)~\cite{gpt-neo} with both Falcon (7B)~\cite{almazrouei2023falcon} and Vicuna (7B)~\cite{vicuna2023}, as well as Llama2 (7B)~\cite{touvron2023llama2}.  We implement these models with the code and checkpoints available from Huggingface library~\citep{wolf2019huggingface}. For Subgraph Retriever~\cite{zhang2022subgraph}, we first retrieve the knowledge graph, and then we incorporate it into the editing LLMs with an editing prompt.

\newpage

\bibliographystyle{ACM-Reference-Format}
\balance
\bibliography{modeledit}

\end{document}